\documentclass[twoside,leqno,twocolumn]{article}

\usepackage[letterpaper]{geometry}

\usepackage{graphicx}
\usepackage{booktabs} 
\usepackage{amsmath}
\DeclareMathOperator*{\argmin}{argmin}
\usepackage{balance}
\usepackage{lineno}
\usepackage{microtype}
\usepackage{caption} 
\usepackage{multirow}
\PassOptionsToPackage{hyphens}{url}
\usepackage{hyperref}
\providecommand{\keywords}[1]
{
  \textbf{\textit{Keywords---}} #1
}

\usepackage{ltexpprt}

\begin{document}

\title{\Large Online Fashion Commerce: Modelling Customer Promise Date\thanks{Supported by GSF grants ABC123, DEF456, and GHI789.}}
\author{Preethi V\thanks{Myntra Designs, India.}
\and Nachiappan Sundaram\thanks{Ericsson, India.}
\and Ravindra Babu Tallamraju\thanks{Myntra Designs, India.}}

\date{}

\maketitle





\fancyfoot[R]{\scriptsize{Copyright \textcopyright\ 2021\\
Copyright retained by principal author's organization}}


\begin{abstract} \small\baselineskip=9pt In the e-commerce space, accurate prediction of delivery dates plays a major role in customer experience as well as in optimizing the supply chain operations. Predicting a date later than the actual delivery date might sometimes result in the customer not placing the order (lost sales) while promising a date earlier than the actual delivery date would lead to a bad customer experience and consequent customer churn. In this paper, we present a machine learning-based approach for penalizing incorrect predictions differently using non-conventional loss functions, while working under various uncertainties involved in making successful deliveries such as traffic disruptions, weather conditions, supply chain, and logistics. We examine statistical, deep learning, and conventional machine learning approaches, and we propose an approach that outperformed the pre-existing rule-based models. The proposed model is deployed internally for Fashion e-Commerce and is operational.\end{abstract}


\section{Introduction.}

The outbound supply chain consists of various stages such as procurement from the vendor, warehouse processing, linehaul shipping, and delivery from the last-mile center. The supply chain and logistics-related findings from PwC\cite{RefWorks:1} show that 39\% of the customers want their goods delivered within 2 days, while 50\% consider 3-5 days as acceptable. However, it is important to have better predictability over speed\cite{RefWorks:1}. Predicting delivery dates accurately at the time of order placement is one of the biggest challenges in the e-commerce space, given the uncertainties involved in each of the supply-chain legs. Once an order is placed, a promise date for delivery is given to a customer based on the expected time taken at every leg. The uncertainties at various stages include traffic, weather conditions, curfew, etc. The additional factors that can be modeled include delays due to holidays, weekends, pendency from an increased load, office location deliveries on weekends. A shipment delivered after a promised date leads to a breach resulting in a bad customer experience. 

In a large developing country like India, building a strong supply chain network capable enough to reach all remote parts of the country on time is a challenge due to limited infrastructure availability in certain parts of the country, unexpected traffic, road blockages, weather conditions, and variable holiday patterns. PwC study showed that retailers are moving from large distribution centers in the heartland to smaller ones close to population centers\cite{RefWorks:1}. Myntra has also strategically located the last-mile centers based on demand concentration, thereby ensuring that the majority of delivery requirements are met by myntra logistics (ML). However, support from third-party logistics (3pl) is imperative to reach the remote parts of the country and places where the demand is less. This adds to the uncertainties, as we do not have clear visibility into the 3pl partners' network in terms of transit hub\footnote{A hub is the distribution center where shipments are collected, sorted, and reloaded.} and last-mile center locations, manpower, and their load from other vendors. There are also challenges arising from limited data availability for low-demand regions and operational variability during fixed and flexible holidays, and weekends at the pick-up, transit, and last-mile locations.
 
In the fashion industry, since it is necessary to have a variety of products to meet a varied set of customers with differing needs, the individual stock-keeping-unit (SKU) depth (stock available for each item) tends to be low. The demand needs to be met by a small number of warehouses where goods are stored as well as processed and also by fulfillment centers to cover regional demand. Strategically distributing the SKUs across them is challenging. Thus, there might be cases where the fulfillment cannot be done from the nearest fulfillment center or warehouse. This leads to a longer linehaul and subsequently more uncertainties resulting from traffic congestion, hub wait times, and regional holidays in transit hubs. 

Promised delivery times are also affected by hard-to-decipher customer addresses, as the addresses may not follow a predefined structure. Incomplete addresses, spell variations, and avoidable additional text such as directions to reach the address, times of their availability pose challenges in automated address clustering models\cite{10.1145/2837689.2837696} designed to ease delivery. Thus quick and successful delivery depends on address validity.

Predicting delivery times during high revenue days (HRD) is another big challenge, as the variability in delivery times would be relatively higher and the shipping cutoffs at the transport and dispatch hubs may not be adhered to. The prediction has to be made based on past HRD events, which typically happen only a couple of times in a year. There would be changes happening in the supply chain since past events' occurrence, for instance, opening, merging, and closing of delivery centers, dispatch hubs, process changes, change in revenue targets, etc. Thus, planning data such as volume and capacity plans at various supply chain touchpoints play a major role in predicting delivery time accurately in case of HRD. Delivery time accuracy is thus also influenced by the accuracy of plan numbers.

In this paper, we briefly describe how accurately predicted times could help in examination of unexpected delays and processing rate changes thus enabling optimisation of the supply chain operations. We compare time series, deep learning and machine learning-based approaches for the current problem. We also examine asymmetric loss functions to meet the business constraint of keeping breaches under control as the cost of a breach is higher than the cost of promising a date much later than the actual delivery date. We also demonstrate the effectiveness of this solution for a large-scale Indian e-commerce company.

In the next section, we briefly describe research work related to the current problem. Section 3 contains problem description and challenges. Section 4 discusses various methodologies applied to solve the problem and concludes on the proposed solution. We present the results of various modeling approaches and real-world deployments in Section 5 and, conclusions and further work in Section 6.

\section{Related Work.}
In this section, we briefly review the existing work on delivery time prediction. Ravindra Babu, et al.\cite{10.1145/2837689.2837696} describe a classification approach to assign addresses to area-codes, based on the address-text. They describe the impact of unstructured, incorrect, or incomplete addresses on the accuracy of classification. Such addresses can impact the on-time delivery of products, especially in India, where addresses do not follow a regular format like in the Western-world. Jeff Berman\cite{RefWorks:1}, discusses customer expectations. Customers have become accustomed to getting products as and when they want them, which creates challenges for the supply chain. Faster deliveries need a high degree of alignment at the supplier, storage, transport, and manpower levels, which can be expensive. Predictions from models, whether rule-based or data-driven, can be erroneous. The predicted delivery date can fall before or after the actual delivery date. The former error is more detrimental for the company (late deliveries) and should be penalized more as compared to the latter error. Models that we build should capture this asymmetry\cite{christoffersen1996further}. 

In \cite{RefWorks:2}, authors talk about predicting lead times for made-to-order industrial-grade equipment. Here, numerous categorical variables affecting times were reduced using the PCA technique. The random forest model gave the best accuracy. In \cite{jonquais2019predicting}, shipping time between factory to port in two continents is predicted using the Random forest model by taking into account port congestion and a major holiday. In this study, it is shown that as the shipment gets closer to the destination, MAE reduces and random forest performs similar to traditional approaches. In almost all the problems, the value lies in the ability to make correct predictions right at the origin time, as customers would not prefer to see updated times at every stage. In \cite{liu2018predicting}, the authors predict purchase order fulfillment times using random forest and quantile regression forests. In this paper, dimension reduction technique has been applied to identify the most influential levels of categorical variables similar to \cite{RefWorks:2}. This paper explores tree-based models, time series and deep-learning based approaches to model delivery times end-to-end, while also demonstrating the use of custom loss functions and feedback-based breach control model to handle asymmetry.

\section{Problem Description and Challenges.}
The objective of the current problem is to accurately predict delivery dates at the time of order placement. Once an order is placed, the item has to traverse through various touchpoints in the supply chain before reaching the customer's location.

The order processing starts from getting the goods from the vendors in case items are not available in the warehouse. The order fulfillment time from the vendor would vary based on the item availability, vendor processing speed, transportation time and related delays, use of co-loader vehicles, and weekend delays due to reduced manpower. It would also depend on the designated times at which the purchase order could be issued and goods can be picked up in a day. Further processing is done at the warehouse. 

The order processing starts at the warehouse for items available in the warehouse and there is no vendor processing leg involved for such items. It broadly involves the activities of picking, consolidation, and packing. An uncertain delay is introduced because of wait times for multi-item packets, cascading effect resulting from occasional choking in one or more previous stages, pendencies due to warehouse closure, delays due to working-shift changes, variable manpower during weekends and flexible holidays, etc. 


\begin{figure}[h]
\centering
    \includegraphics[height=3cm, width=70mm]{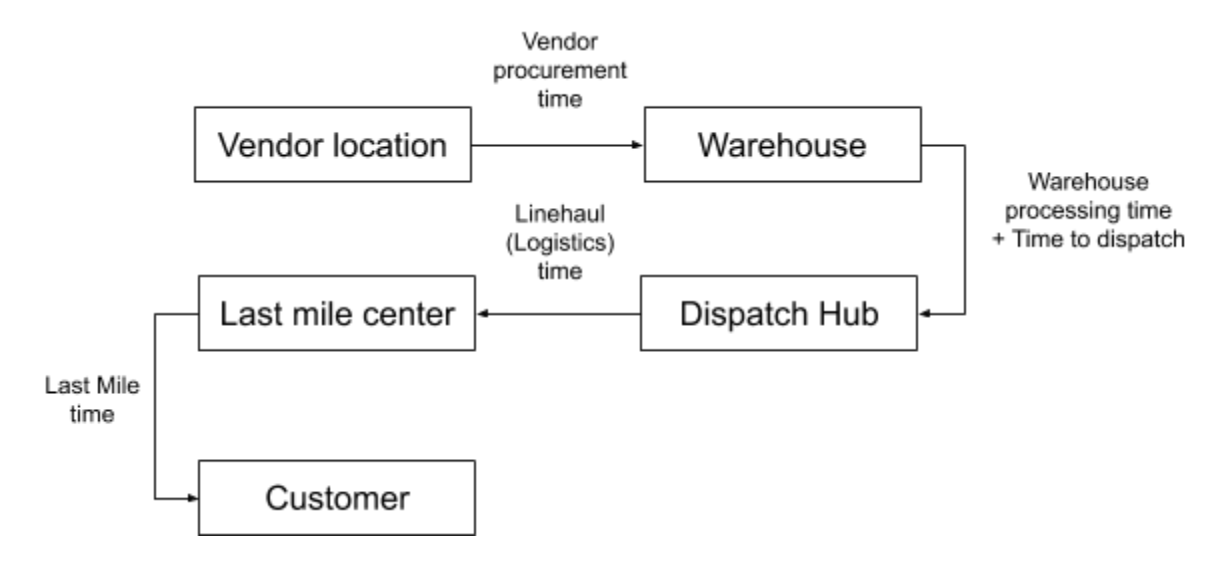}
\caption{Order processing stages}
    \label{fig:orderproc}
    \vspace{-2mm}
\end{figure}

Subsequently, a shipment traverses through a sequence of hubs where transit time is influenced by traffic congestion, weather conditions, hub wait times until shipping cutoff, regional holidays in transit hubs, etc.

When the packet reaches the last-mile hub, one of the factors causing delay is variable manpower on holidays and weekends. Each holiday duration depends on the nature of the holiday such as religious festivals. The delay in packet delivery is a function of current pendency, available manpower, customer availability, and deliverability of an address. However, it is difficult to localize which packets would be delivered late as it depends on customer availability and delivery re-attempt. The order processing stages are summarized in Figure \ref{fig:orderproc}.

\section{Proposed Solution.}

We propose to solve this problem in three stages. The following is the set of models built for predicting the processing time involved in various supply-chain legs:
\vspace{0.1cm} \newline \textbf{Vendor procurement model:} The time taken from order placement to shipping is predicted using this model. This is applicable only for items not stored in the warehouse. This time is modeled using LightGBM\cite{ke2017lightgbm} and explained in Section 4.8.
\newline \textbf{Warehouse model:} This model predicts the time taken from order placement to shipping only for items stored in the warehouse. This time is modeled using time series (fbprophet\cite{taylor2018forecasting}) and LightGBM as described in Section 4.6.2.
\newline \textbf{Shipping time model:} This model predicts the time taken to reach customer location after shipping referred to as the shipping time. This constitutes the linehaul and last-mile times as indicated in Figure \ref{fig:orderproc}. The shipping time for ML during Business As Usual(BAU) days is predicted using tree-based models, time-series, and deep learning-based approaches. Tree-based models, namely, LightGBM, Random forest, and XGBoost are described in Section 4.2. The time series modeling approach using fbprophet is described in Section 4.6.1. The Deep-learning based approach (LSTM) is explained in Section 4.7. The shipping time for HRD is modeled using LightGBM and explained in Section 4.3. Shipping time for 3pl partners is modeled using LightGBM and explained in Section 4.4.\newline The various methods used to control breaches such as asymmetric custom loss function, quantile regression, and feedback-based breach control model are discussed in Section 4.5.

At the time of order placement, we propose to give the delivery date to the customer by adding the outputs from the Warehouse model and Shipping time model to the order placement time for items stored in the warehouse. In case of items not stored in the warehouse, the outputs from the Vendor procurement model and Shipping time model will be added to the order placement time.

In this section, we initially present the rule-based model which was operational before the proposed model. Subsequently, we discuss tree-based models for predicting the shipping time of ML and 3pl. The modeling approach for HRD is also described. We then describe ways to control breaches. We discuss time series models to predict shipping time and warehouse time. Subsequently, we present the LSTM model to predict shipping time and the tree-based model to predict vendor procurement time. Then, we briefly discuss applications in supply-chain optimization.

\subsection{Rule-based model to predict delivery time.}
In the past, a rule-based model was used to predict delivery times. At the level of vendor-warehouse, vendor processing times are configured. Similarly, at a warehouse level, processing times are configured for each of the warehouse legs. In the case of linehaul, the hop-to-hop times are configured and static last-mile center processing time is added. Each of these times is mapped to the nearest shipping cutoffs wherever appropriate. Fixed additional processing times are added for weekends and holidays. Thus, rule-based model doesn't adapt based on recent performance changes and is designed based on heuristics.

\subsection{Tree-based models to predict shipping time for Business As Usual(BAU) days.}
\subsubsection{Model choice.}
We examined LightGBM\cite{ke2017lightgbm}. It has the ability to represent categorical features with a large set of levels. The current problem involves categorical features with up to hundreds of levels, which blow up the feature space when one-hot encoded. The selection of optimal split points in tree-based models would thus consume a large amount of time and we may not be able to meet the latency requirements. LightGBM uses exclusive feature bundling to optimally bundle mutually-exclusive features. This works especially when the feature space is sparse as in the case of one-hot encoded features.

Time consumed for the selection of optimal split points is also directly proportional to the number of training examples. Given the huge number of orders placed in the last few weeks before prediction, it is necessary to reduce the number of training examples without hurting accuracy. Gradient-based one-side sampling\cite{ke2017lightgbm} achieves this by randomly-sampling instances with smaller gradients while not changing the data distribution (using weights). Thus, the instances with larger gradients are majorly used to determine the split points. Random forest and XGBoost models were also built on the dataset. 

\subsubsection{Model features.}
Following are the broad list of features considered to take into account various factors affecting shipping time:
\vspace{0.1cm} \newline \textbf{Historical times and pendencies}: The pendency in the linehaul and last-mile plays a major role in determining accurate times. The performance would vary for each logistic lane and last-mile center, which is modeled using statistical measures. Pendency is modeled using the recent inflow and outflow of shipments.
\newline \textbf{Geographical features}: The geographical location level features of customer addresses such as city tier, address type (home or office), area to name a few play an important role in determining delivery time. This helps the model to learn the patterns in time by aggregating geographical locations and thus generate near-accurate predictions for locations with limited training data.
\newline \textbf{Load features}: The expected load at various touchpoints plays a key role in determining delays. The on-ground order and ship volumes from the past days play a major role in determining pendencies at all further stages in the supply chain process.
\newline \textbf{Manpower features}: Various manpower types available at the last-mile center are modeled individually to capture productivity and process differences.
\newline \textbf{Last-mile performance features}: Arrival, delivery volumes and performance at the last-mile center on the past days are modeled taking into account cumulative pendency from past days.
\newline \textbf{Holiday type and seasonality features}: The transit hub and last-mile center holidays are captured using these features. Sudden change in the manpower, arrival, and delivery volumes at the last-mile center might occur due to full or partial closures on holidays/weekends or varied outflow on some weekdays. Hence, the holiday and weekday types are included as features so that the model can effectively learn the relationships between variables.


\subsubsection{Model tuning.}
The examples are weighted using a decaying weighting function as the most recent supply chain operations have a greater impact on current delivery times. As the LightGBM model tends to overfit, we tuned various parameters as indicated in Table \ref{table:Table_1}. The validation set has to be carefully chosen in a time series context as it should be similar to the test set in terms of order day and also meet the recency criteria in this case.

\begin{table}[h!]
\centering
\caption{Parameter tuning for LightGBM}
\small
\begin{tabular}{|c|c|c|c|}
\hline
{Metric}            & {\begin{tabular}[c]{@{}c@{}}Final \\ Value\end{tabular}} & {\begin{tabular}[c]{@{}c@{}}Default \\ Value\end{tabular}} & {\begin{tabular}[c]{@{}c@{}}Parameter value \\ list\end{tabular}} \\ \hline
{\begin{tabular}[c]{@{}c@{}}Boosting \\ iterations\end{tabular}}     & 1000                                                            & -                                                                 & \begin{tabular}[c]{@{}c@{}}{[}1000, 1500, 2000, 2500{]}\end{tabular}  \\ \hline
{\begin{tabular}[c]{@{}c@{}}Tree \\ depth\end{tabular}}        & -1                                                              & -1                                                                & {[}4, 5, 8, -1{]}                                                        \\ \hline
{\begin{tabular}[c]{@{}c@{}}Number \\ of leaves\end{tabular}}       & 15                                                              & 31                                                                & {[}15, 31, 63, 127{]}                                                    \\ \hline
{\begin{tabular}[c]{@{}c@{}}Data \\ fraction\end{tabular}}         & 0.6                                                             & 1.0                                                               & {[}0.6, 0.7, 0.8, 1.0{]}                                                 \\ \hline
{\begin{tabular}[c]{@{}c@{}}Feature \\ fraction\end{tabular}} & 0.6                                                             & 1.0                                                               & {[}0.6, 0.7, 0.8, 1.0{]}                                                 \\ \hline
\end{tabular}
\bigskip
\label{table:Table_1}
\vspace{-6mm}
\end{table}

\subsubsection{Holiday and weekend handling times.}
Weekends and holidays play a major role in determining delivery times. Every last-mile center behaves differently on different holidays based on the holiday type(fixed or flexible), the intensity of the holiday in that region, and the adjacency of the holiday to a weekend. Weekend behavior also differs for every last-mile center. The effect of holiday or weekend is modeled by selecting the most recent similar holidays(based on absenteeism rates) or weekends. The proxy holidays/weekends are then compared with BAU days that lie close to them. The chosen BAU days are subjected to outlier treatment. Statistical measures that are robust against outliers are used to compare and subsequently derive holiday and weekend handling times.

\subsection{Tree-based model to predict shipping time for HRD.}
Delivery times of orders placed on both HRD and post-HRD days are influenced by huge pendencies at various supply chain touchpoints. Accurate delivery time prediction lies in the ability to accurately project pendencies into the future, apart from modeling other factors such as manpower, capacity, volume, holiday, weekend effects, etc.
\subsubsection{Modeling approach.}
LightGBM model is trained using sale and post-sale data on the proxy and pre-proxy days taken from the recent sale event. The proxy day is chosen based on the number of days elapsed from the start of the sale until the prediction date. The planning data for the upcoming event is fed as input to the model along with the on-ground performance features derived from the past days of the sale. This enables the model to correct itself based on recent data in case of deviation from the plan.
\subsubsection{Model features.} 
Apart from the features considered for a BAU day, the following are the additional set of features considered to take into account various factors affecting shipping time during HRD
\vspace{0.1cm}\newline \textbf{Planned volume and capacity features}: The planned ship volume is modeled by combining sources appropriately to project landings at subsequent supply chain touchpoints. The planned last-mile capacity is indicative of the processing capability or cumulative outflow.
\newline \textbf{Pendency features}: Three major components contribute to the delivery time once shipments arrive at the delivery center – arrival volume on the predicted landing date, pendency accumulated from the days prior to the predicted landing date, and delivery center outflow modeled using capacity plan data. These features are modeled using three different data sources or methods. In addition to modeling using plan data, most recent delivery center performance is captured as the event progresses and arrival volume and pendency at the delivery center is projected on the landing day using recent order and ship volume data, in order to combat any deviation from the plan.

\subsection{Tree-based model to predict 3pl shipping time.}
LightGBM is used to model the shipping time for 3p courier partners. The holiday and weekend behavior at the pickup location as well as at the 3pl last-mile center was handled using the methodology described earlier based on the regional holiday calendar of the courier partners. However, some of the factors influencing shipping time as described above cannot be explicitly captured in the case of 3pl as we are not aware of the last-mile center locations, corresponding pincode mapping and arrival times. Pincode decoding is used to predict times for low-demand regions where limited or no recent data is available.

\subsection{Methods to control breaches.}

\subsubsection{Asymmetric loss functions.}
Most machine learning models operate by default on loss functions that are minimized by central tendencies, either the mean or the median. The sum of squared residuals has its minimum at the mean, while the sum of absolute residuals is minimized by the median. Here, the goal is to penalize underestimations, that is when the predicted time is less than the actual time. A penalization factor is applied for the negative residuals on the default MSE (mean square error) loss\cite{christoffersen1996further}. The minimization of this loss function is indicated in equation(\ref{eqn:01}).

\begin{align}\label{eqn:01}
&e_i(\beta) = \begin{cases}
  \alpha (y_i - f(\beta, X_i)), & \text{if } y_i > f(\beta, X_i) \nonumber \\
  (y_i - f(\beta, X_i)) & \text{otherwise}
 \end{cases}\\
&\beta^{*} = \argmin_{\beta} \left( \frac{1}{n}\sum_i {e_i}^2(\beta) \right)
\end{align}

The problem in using any asymmetric loss function of the above type is that predictions are still aimed at the expected value. We need to get at a prediction that is greater than the true value by a specified probability, that is, the breach cut-off. Multiplying the prediction obtained using MSE loss by a factor does not help us get at the desired result, as the central tendencies and higher percentiles are not linearly related.

\subsubsection{Quantile regression.}
To maintain breaches below a specified percentage, we need to be able to predict the quantile of the probability distribution of future times. One way would be to get the full predictive densities and pick the desired quantile. However, the more powerful algorithms only give point predictions.

\begin{figure}[h]
\centering
    \includegraphics[width=60mm,height=30mm]{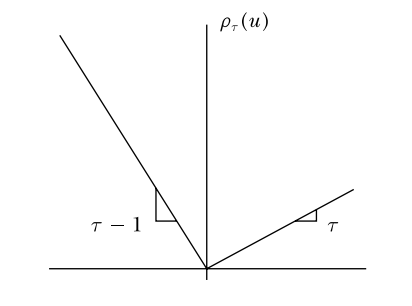}
\caption{Quantile loss function\cite{koenker2001quantile}}
    \label{fig:quantilelossfinal}
    \vspace{-2mm}
\end{figure}

Quantile regression\cite{koenker2001quantile} is a modeling approach that aims to predict the percentiles by optimizing an objective that has its minimum at the specified quantile. The quantile loss function is the sum of asymmetrically weighted absolute residuals. The minimization of this loss function is shown in equation(\ref{eqn:02}) and the loss function is visualized in Figure \ref{fig:quantilelossfinal}. The solution to the below equation is the estimate of the conditional quantile function. LightGBM supports quantile regression which can be specified using the objective parameter.

\begin{equation}\label{eqn:02}
\beta^{*} = \argmin_{\beta} \sum_i \rho _{\tau }(y_{i}-f ({X_{i}},\beta ))
\end{equation}

\subsubsection{Feedback-based Breach control model design.}
In this approach, we built a feedback mechanism to correct the output from the LightGBM model (with default MSE loss function). We feed the predictions made in the last few days as input to a breach control model, along with the corresponding weekend, holiday handling times, holiday, seasonal features, and statistical measure-based features used to capture variations in linehaul and last-mile time. The target is constructed as a weighted combination of linehaul standard deviation, the mean and standard deviation of packet inflow and outflow at the last-mile. These weights are tuned based on past model performance to get the breaches below the specified percentage at a delivery date level. Model training is done on this dataset with ideal additional times as the target to predict the optimal time required to get breaches just below the desired level.\newline
The output from this model is added to the predictions made by the LightGBM model.

\subsection{Time-series modeling approach.}

\subsubsection{Time series model to predict shipping time.}
\begin{figure}[h]
    \includegraphics[width=80mm,height=4cm]{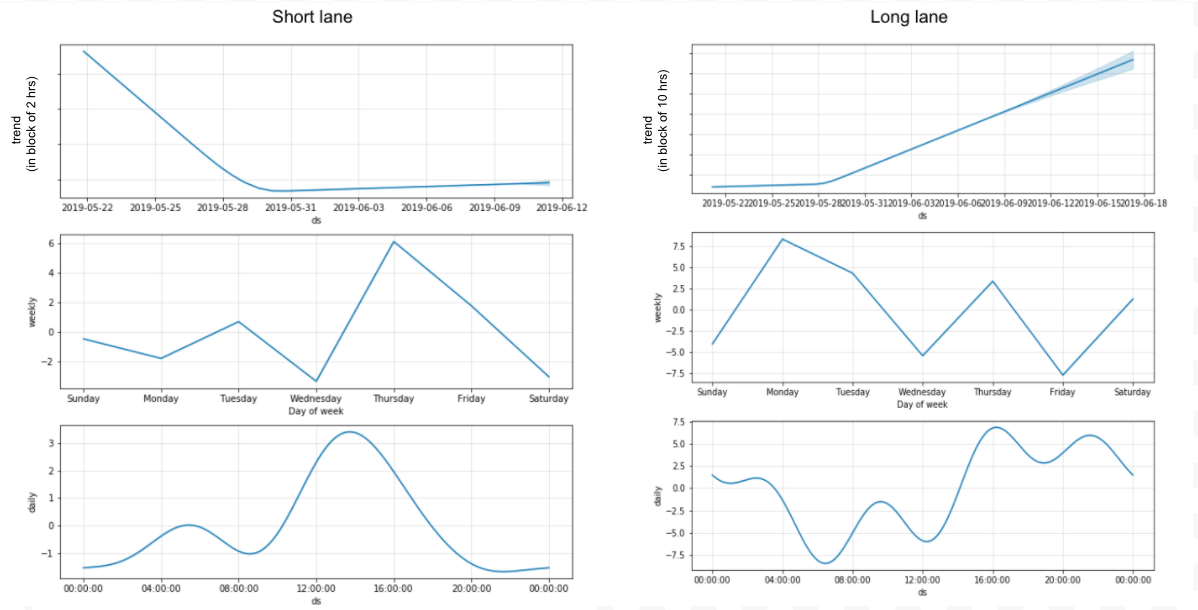}
\caption{Trend, weekly and daily seasonality plots for short and long lanes}
    \label{fig:timeseriesshiptattrend}
    \vspace{-2mm}
\end{figure}

\begin{figure}[h]
    \includegraphics[width=80mm,height=3cm]{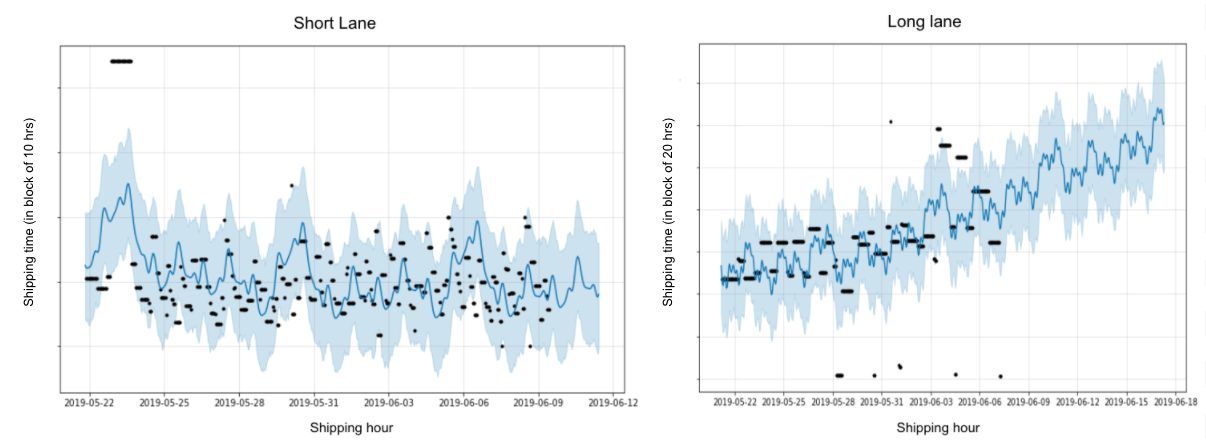}
\caption{Fbprophet model fitting for a short versus long lane.}
    \label{fig:timeseriesshiptatfinalll}
    \vspace{-2mm}
\end{figure}

In order to model time-dependencies, we used the time series approach. The Fbprophet model handles trend, seasonality, and holiday effects that occur on irregular schedules as indicated in equation(\ref{eqn:03}). It can be seen from the weekly seasonality plot shown in Figure \ref{fig:timeseriesshiptattrend}\footnote{The y-axis numbers are hidden to not reveal internal business numbers.} that the shipping times are higher for orders shipped on Thursday, in case of short lanes, and for orders shipped on Monday and Tuesday, in case of longer lanes. This pattern is seen as these shipments would typically land in the last-mile center on a weekend. The \textit{changepoint prior scale} parameter helps determine how flexible the trend should be. This parameter was tuned to avoid overfitting. Cap and floor parameters were set to specify the range within which the predictions can lie. Regional holiday list was inputted. The data were sampled based on hourly frequency. Predictions were made at an hourly level and the upper end of the prediction interval was used as the final result to control breaches. The model fitting is shown in Figure \ref{fig:timeseriesshiptatfinalll}\textsuperscript{2}. Black dots indicate training data/known target values. The blue line indicates the predictions made on the train and test data. The blue shaded region indicates the interval for the predictions.

\begin{equation}\label{eqn:03}
y(t) = g(t) + s(t) + h(t) + \epsilon_{t}
\end{equation}

\subsubsection{Time series model to predict warehouse time.}
\begin{figure}[h]
\centering
    \includegraphics[height=3.5cm,width=70mm]{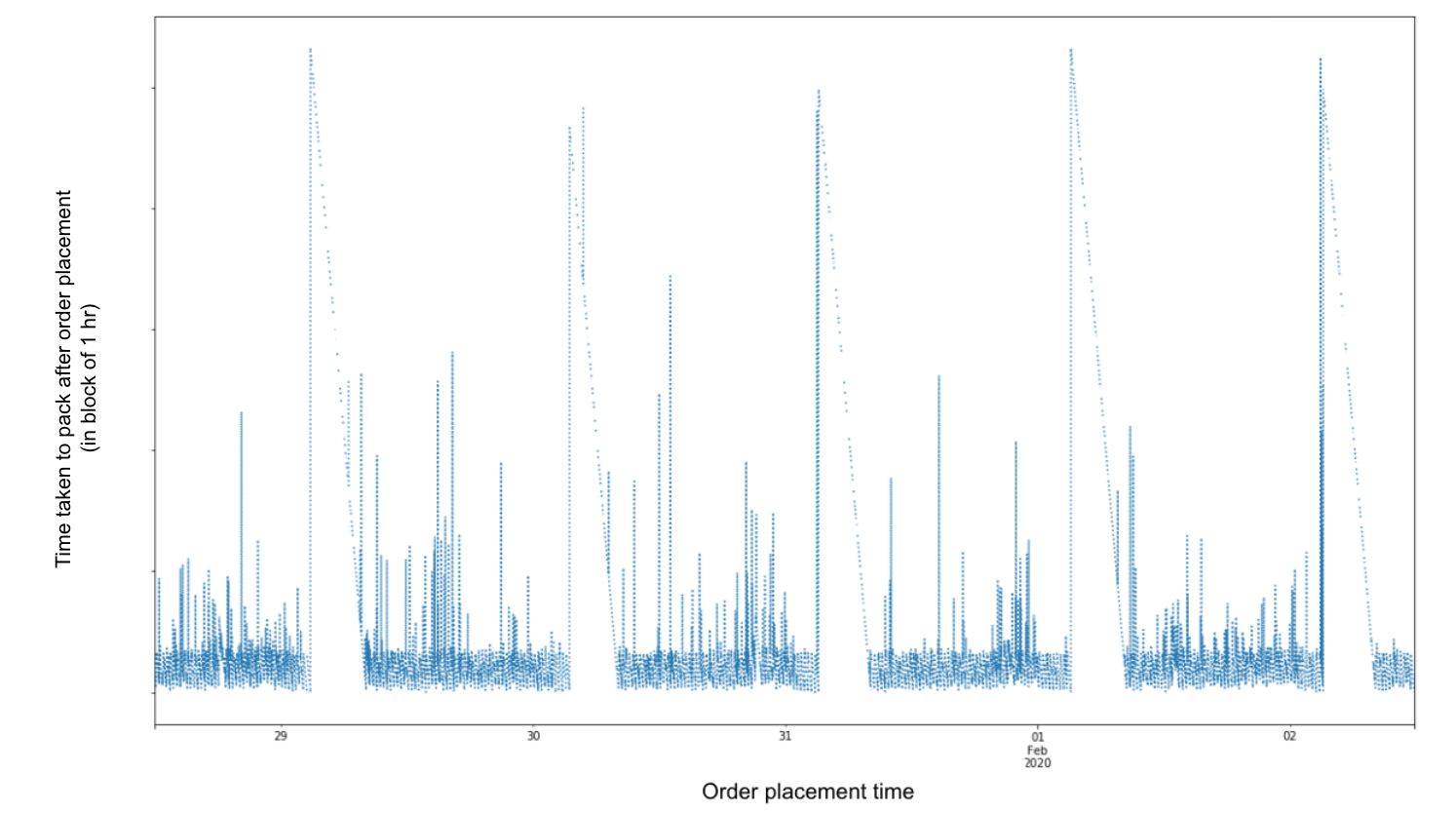}
\caption{Time series for Warehouse processing leg (Order placement to pick)}
    \label{fig:otop}
    \vspace{-2mm}
\end{figure}

Fbprophet model was used to predict the warehouse processing time. The time taken from order placement to pick is shown in Figure \ref{fig:otop}\textsuperscript{2}. The periodic patterns are seen due to shift timings. Models were built separately for single-item and multi-item orders as multi-item packets have an additional wait time involved to consolidate all items. The seasonal effects by the hour and day of the week are captured by the fbprophet model as the model is built on the past few weeks of data. The time-series data was resampled based on minute-level frequency. Re-training is done every hour to capture the recent changes and predictions are made at a minute level for the next hour.

As the warehouse processing time is small even when compared to a short lane, the model was able to capture recent changes effectively and provide accurate real-time predictions. A LightGBM model was also built using the projected load, multi-item based, seasonal and performance-based features.

\subsection{Deep-learning based approach.}
\subsubsection{LSTM model to predict shipping time.}
\begin{figure}[h]
    \includegraphics[width=80mm]{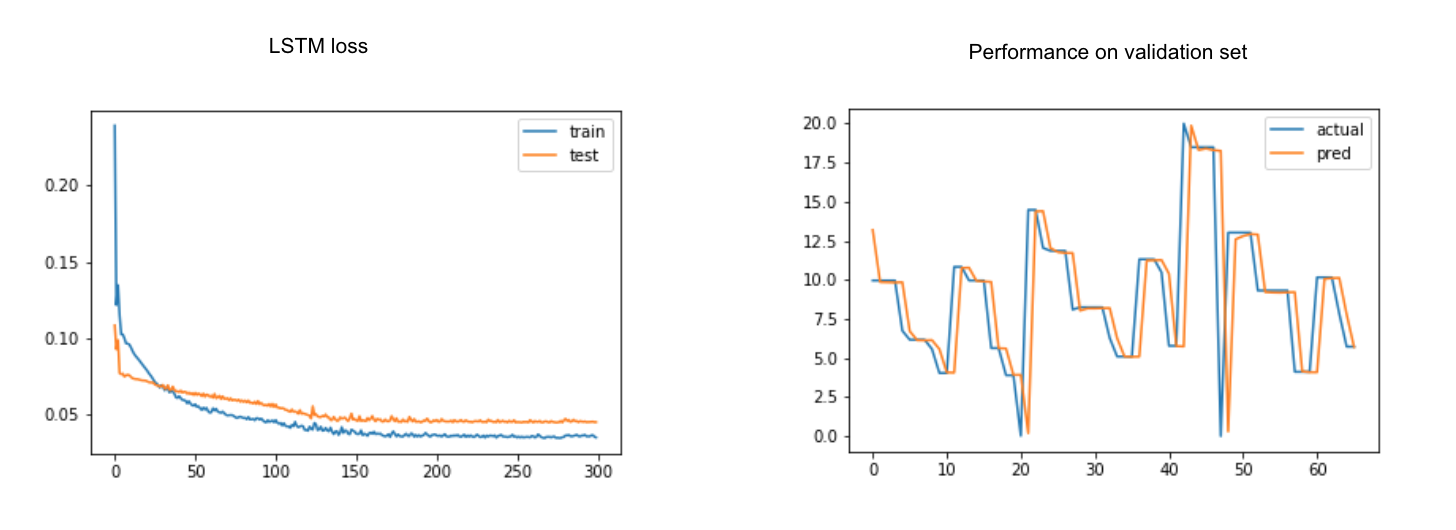}
\caption{LSTM loss and performance on the validation set}
    \label{fig:lstmlossvalset}
    \vspace{-2mm}
\end{figure}

LSTM is a powerful recurrent neural network that handles sequence dependence among input variables. The network was designed to have 7 LSTM blocks in the hidden layer to capture seasonality. This LSTM network was trained for 300 epochs using a batch size of 100. The lookback value was chosen to cover few weeks after tuning. The other relevant features affecting shipping time were also included. From Figure \ref{fig:lstmlossvalset}, we can see that the model is able to fit the dataset well. 

\subsection{Tree-based model to predict vendor procurement time.}
LightGBM was used to model the time taken to procure the items not available in the warehouse from the vendor and subsequently ship the items. Several features were built including vendor type, features based on designated hours at which purchase order is generated and the times at which pick up happens, location and route-based features, seasonality features, the maximum time allowed to process a purchase order, and pick up in coloader vehicles.

\subsection{Supply-chain optimization.}
Modeling delivery times accurately not only helps in improving customer experience and reducing customer churn but also in providing business insights to optimize the supply chain operations. When the on-ground processing time shows deviation from the predicted time, it can be used to flag and subsequently correct unexpected pendencies or delays at an early stage. Analysis on last-mile center processing speed showed that making all delivery attempts within a fixed time duration post arrival would help in reducing breach. This further enables us to provide more accurate times.

\section{Experimental Results.}
In this section, we discuss the experimental results, where we considered a large dataset for training and evaluation. We examine various models such as variants of LightGBM, XGBoost, Random forest, LSTM, and time series models, and compare the performance with the current baseline of rule-based models. We also present results for HRD and different linehauls. We summarise the results for various supply chain legs such as Warehouse, Shipping(3pl), and Vendor procurement.

\subsection{Data.}
The training data constitutes deliveries made in the last few weeks prior to the prediction date amounting to more than a \textbf{million training examples} and the testing is done on the orders placed on the subsequent day.
\subsection{Evaluation metrics.}
Accuracy is measured as the percentage of total packets ordered in a day that get delivered on the same day(0) as the promise date or the previous day(-1) and is denoted as Accuracy(0 to -1). Breaches refer to the percentage of total packets ordered in a day that get delivered after the promised date. These are delivery date-based metrics measured at an order-day level.

\subsection{Results.}
Table \ref{table:Table_2} shows order-day level metrics averaged over a period of one week in the month of September, thus covering weekend effects. The results cover all possible lanes amounting to around thousands and correspond to only items stored in the warehouse. The models mentioned in the table were used to predict shipping time. The time from order placement to shipping is taken from the rule-based model.

\begin{table}[h!]
\centering
\caption{Comparison of model performance covering all possible lanes on Business As Usual(BAU) days}
\small
\begin{tabular}{|c|c|c|}
\hline
\textbf{Model}                                                                                      & \textbf{\begin{tabular}[c]{@{}c@{}}Accuracy\end{tabular}} & \textbf{Breach} \\ \hline
\textbf{\begin{tabular}[c]{@{}c@{}}LightGBM (Quantile  \\regression)\end{tabular}}    & \textbf{68.10\%}                                                      & \textbf{4.68\%}   \\ \hline
\begin{tabular}[c]{@{}c@{}}LightGBM (Asymmetric \\loss function)\end{tabular} & 61.54\%                                                               & 5.50\%            \\ \hline
\begin{tabular}[c]{@{}c@{}}LightGBM \\(with breach control model)\end{tabular}       & 58.30\%                                                               & 3.90\%            \\ \hline
\begin{tabular}[c]{@{}c@{}}XGBoost \\(with breach control model)\end{tabular}        & 58.13\%                                                               & 3.82\%            \\ \hline
\begin{tabular}[c]{@{}c@{}}LSTM \\(with breach control model)\end{tabular}          & 57.25\%                                                               & 6.20\%            \\ \hline
\begin{tabular}[c]{@{}c@{}}Random forest \\(with breach control model)\end{tabular}  & 55.45\%                                                               & 3.85\%            \\ \hline
\begin{tabular}[c]{@{}c@{}}Time series model \\(fbprophet)\end{tabular}                                                                        & 40.30\%                                                               & 2.74\%            \\ \hline
\begin{tabular}[c]{@{}c@{}}Rule-based model \\(Baseline)\end{tabular}                                                                          & 36.56\%                                                               & 5.63\%            \\ \hline
\end{tabular}
\bigskip
\label{table:Table_2}
\vspace{-6mm}
\end{table}


It can be seen from Table \ref{table:Table_2} that the quantile regression approach achieved the highest accuracies, almost doubling the accuracy achieved using the baseline model, while also reducing the breaches. LightGBM trained with asymmetric loss function resulted in relatively higher breach, next to the rule-based model. Though the penalization factor was tuned, breaches could not be deterministically controlled using this approach. LightGBM in conjunction with the breach control model has achieved relatively a lower breach while taking a slight hit on accuracy. The breach control model is trained based on breaches at a delivery date level to meet recency criteria. For every order date, the factors are tuned to meet breach constraints at all the delivery dates when those shipments landed. The factor tuning is thus also influenced by performance on other order dates. Therefore, this method gives a more conservative breach result. LightGBM model accuracy is slightly higher than that achieved using XGBoost, while model training with LightGBM was 8x times faster. The Random forest model provided lower accuracies when compared to both LightGBM and XGBoost models.

\begin{table}[h!]
\centering
\caption{Comparison of model performance covering all possible lanes on HRD and post-HRD days}
\small
\begin{tabular}{|c|c|c|}
\hline
\textbf{Model}                                                                                      & \textbf{\begin{tabular}[c]{@{}c@{}}Accuracy\end{tabular}} & \textbf{Breach} \\ \hline
\begin{tabular}[c]{@{}c@{}}LightGBM - HRD\end{tabular}    & 65.15\%                                                      & 7.86\%   \\ \hline
\begin{tabular}[c]{@{}c@{}}LightGBM - Post-HRD\end{tabular} & 74.58\%                                                               & 4.82\%            \\ \hline
\begin{tabular}[c]{@{}c@{}}Rule-based model - HRD\end{tabular}                                                                        & 44.23\%                                                               & 5.44\%            \\ \hline
\begin{tabular}[c]{@{}c@{}}Rule-based model - Post-HRD\end{tabular}                                                                         & 54.48\%                                                               & 3.06\%            \\ \hline
\end{tabular}
\bigskip
\label{table:Table_5}
\vspace{-6mm}
\end{table}

Table \ref{table:Table_5} shows order-day level metrics averaged over the HRD and post-HRD days in the month of December. Accuracy(0 to -2) is the performance metric measured in the case of HRD as predictability is more challenging compared to BAU days. The results cover all possible lanes and correspond to both items stored in the warehouse as well as those procured from the vendor. The models mentioned in the table were used to predict shipping time. The time from order placement to ship is assumed to be known to indicate the performance of the shipping time model without considering the inaccuracies resulting from the pre-ship leg. Note that since the breaches at the warehouse side are controlled during HRD, the breaches at the time of order placement were still under control despite a slightly higher breach percentage on the shipping leg. It can be seen from Table 3 that the best LightGBM implementation achieved an improvement of more than 20 percentage points over the baseline model in the case of both HRD and post-HRD days.

\begin{table}[h!]
\centering
\caption{Comparison of model performance for a short versus long lane}
\small
\begin{tabular}{|c|c|c|c|c|}
\hline
\multirow{2}{*}{\textbf{Model}} & \multicolumn{2}{c|}{\begin{tabular}[c]{@{}c@{}}Short Lane \\ (Delhi - Gurugram)\end{tabular}} & \multicolumn{2}{c|}{\begin{tabular}[c]{@{}c@{}}Long lane \\ (Delhi - Kolkata)\end{tabular}} \\ \cline{2-5} 
                                & \textbf{\begin{tabular}[c]{@{}c@{}}Accuracy \end{tabular}}   & \textbf{Breach}  & \textbf{\begin{tabular}[c]{@{}c@{}}Accuracy \end{tabular}}  & \textbf{Breach} \\ \hline
\begin{tabular}[c]{@{}c@{}}Time-\\series\end{tabular}                       & 83.55\%                                                                  & 0.52\%             & 50.34\%                                                                 & 2.54\%            \\ \hline
LSTM                            & 83.55\%                                                                  & 0.52\%             & 70.65\%                                                                 & 7.24\%            \\ \hline
\textbf{\begin{tabular}[c]{@{}c@{}}Light-\\GBM\end{tabular}}               & \textbf{83.55\%}                                                         & \textbf{0.43\%}    & \textbf{75.26\%}                                                        & \textbf{5.28\%}   \\ \hline
\end{tabular}
\bigskip
\label{table:Table_3}
\vspace{-6mm}
\end{table}

The comparison of LSTM (in conjunction with breach control model), fbprophet (Time series model), and LightGBM (Quantile regression implementation) models was carried out for a short versus long lane as shown in Table \ref{table:Table_3}. It describes order-day level metrics averaged over a period of one week in the month of September and corresponds to only items stored in the warehouse. LSTM and fbprophet worked on par with LightGBM in the case of shorter lanes. While for longer lanes, the time series model has to predict much farther into the future due to non-availability of recent target values. This can be seen in Figure \ref{fig:timeseriesshiptatfinalll}\textsuperscript{2}. This leads to reduced confidence in the predictions. Fbprophet picked up recent trend changes and made predictions much ahead into the future based on these changes, thus resulting in overfitting. In the supply chain problem, the recent past has an impact only on the immediate future but the effect fades away as we go into the future. The LightGBM model tends to average out the effects while taking into account the potential effect from other factors on the prediction day. In the case of LSTM as well, predictions much ahead into the future are made based on existing predictions. However, it performed much better than the time series model. The results follow a similar pattern across all lanes as shown in Table \ref{table:Table_2}.

\begin{table}[h!]
\centering
\caption{Comparison of Model Performance for the various supply chain legs}
\small
\begin{tabular}{|c|c|c|c|c|}
\hline
\textbf{\begin{tabular}[c]{@{}c@{}}Supply \\ chain leg\end{tabular}}                    & \textbf{Metrics}                                                      & \begin{tabular}[c]{@{}c@{}}Rule-\\ based\end{tabular} & \textbf{\begin{tabular}[c]{@{}c@{}}Time\\ series\end{tabular}} & \textbf{\begin{tabular}[c]{@{}c@{}}Light-\\GBM\end{tabular}} \\ \hline
\multirow{2}{*}{\textbf{\begin{tabular}[c]{@{}c@{}}Ware-\\house\end{tabular}}}                                                     & \textbf{\begin{tabular}[c]{@{}c@{}}Accuracy\end{tabular}} & 82.23\%                                               & \textbf{88.19\%}                                               & 86.20\%           \\ \cline{2-5} 
                                                                                        & \textbf{Breach}                                                     & 3.79\%                                                & \textbf{6.54\%}                                                & 5.32\%            \\ \hline
\multirow{2}{*}{\textbf{\begin{tabular}[c]{@{}c@{}}Shipping\\ (3pl)\end{tabular}}}        & \textbf{\begin{tabular}[c]{@{}c@{}}Accuracy\end{tabular}} & 9.41\%                                                & -                                                              & \textbf{55.52\%}  \\ \cline{2-5} 
                                                                                        & \textbf{Breach}                                                     & 1.24\%                                                & -                                                              & \textbf{3.70\%}   \\ \hline
\multirow{2}{*}{\textbf{\begin{tabular}[c]{@{}c@{}}Vendor -\\ customer\end{tabular}}} & \textbf{\begin{tabular}[c]{@{}c@{}}Accuracy\end{tabular}} & 8.65\%                                                & -                                                              & \textbf{40.84\%}  \\ \cline{2-5} 
                                                                                        & \textbf{Breach}                                                     & 1.34\%                                                & -                                                              & \textbf{5.06\%}   \\ \hline
\end{tabular}
\bigskip
\label{table:Table_4}
\vspace{-6mm}
\end{table}

The results shown in Table \ref{table:Table_4} are also reported at an order-day level by averaging over a period of one week. In the case of 3pl, the models mentioned in the table were used to predict shipping time. The time from order placement to shipping is taken from the rule-based model, similar to the reporting done in Tables \ref{table:Table_2} and \ref{table:Table_3}. 
For the warehouse leg, the accuracy and breaches are reported for the predicted ship date (time from order placement to ship is predicted by the models mentioned in the table). These results correspond to only items stored in the warehouse. The results reported against vendor - customer are accuracies and breaches for the predicted delivery date. These results correspond to only items not stored in the warehouse as the vendor procurement stage is applicable only for such items. The models mentioned in the table are used to predict the time taken from order placement to shipping and the time taken after shipping to reach the customer is taken from the best LightGBM implementation. As the warehouse and vendor procurement models are in the early stages, we have reported the metrics in Tables \ref{table:Table_2} and \ref{table:Table_3} by taking the time until shipping from the rule-based model.

It can be seen that the fbprophet model has resulted in an accuracy improvement of 6 percentage points over the rule-based model. The breaches are slightly higher as no breach control mechanism is applied at the warehouse processing leg, it is compensated for when combined with predicted times of other order processing legs. It also performed slightly better than the LightGBM model in this case, as the recent changes in performance could be captured to give accurate predictions. The LightGBM model for 3pl has achieved a 45\% jump in accuracy while working under the constraints imposed by 3pl data. Again in the case of vendor processing time where we do not have complete control, accuracy has improved by 31\%, with a breach percentage slightly above the cutoff level.

\section{Summary.}
Promise time computation is a challenging problem in e-commerce given uncertainties at various stages such as variability in vendor processing time, consolidation wait time, pendencies and additional uncertainties at the warehouse, traffic disruptions, weather conditions and hub wait times during linehaul, pendency, manpower variation, and customer availability at last-mile, regional holiday and weekend effects at all of the above legs. We propose a solution that integrates all these aspects. The solution is deployed and operational for Myntra. In the paper, we discuss the challenges in detail and provide empirical evidence through data analysis. We have considered a number of models before zeroing on the final solution such as multiple variants of LightGBM, XGboost, Random forest, time series model (fbprophet), and deep learning approaches.  We have carried out elaborate experiments on more than a million patterns and compared a range of approaches covering these models, while also specifying methods to meet the business constraints using \textbf{asymmetric loss functions} and a \textbf{feedback-based breach control model}. The LightGBM implementation of quantile regression gave the best results among the chosen models, almost doubling the accuracy achieved with the baseline model. We have proposed a model for handling high revenue days, where predictability is more challenging due to higher variability in times, deviation from the plan, and process changes. We also built a promise time prediction model for 3pl. As future work, we propose to enhance vendor procurement and warehouse models and comprehensively provide end-to-end predictability.

\bibliographystyle{plain}
\bibliography{name}

\end{document}